\documentclass{article}
\usepackage{spconf,amsmath,graphicx}
\usepackage{subfig}
\graphicspath{{fig/}}
\DeclareGraphicsExtensions{.eps}

\usepackage{booktabs}       % professional-quality tables
\usepackage{amsfonts}       % blackboard math symbols
\usepackage{nicefrac}       % compact symbols for 1/2, etc.
\usepackage{microtype}      % microtypography
\usepackage{amsmath}
\usepackage{graphicx}
\usepackage{subfig}

\title{Character-Level Language Modeling \\ with Hierarchical Recurrent Neural Networks}
\name{Kyuyeon Hwang and Wonyong Sung\thanks{This work was supported in part by the Brain Korea 21 Plus Project and the National Research Foundation of Korea (NRF) grant funded by the Korea government (MSIP) (No.~2015R1A2A1A10056051).}}
\address{Department of Electrical and Computer Engineering \\
Seoul National University \\
1, Gwanak-ro, Gwanak-gu, Seoul, 08826 Korea \\
\texttt{kyuyeon.hwang@gmail.com; wysung@snu.ac.kr}}
\begin{document}
\ninept
\maketitle
\begin{abstract}
Recurrent neural network (RNN) based character-level language models (CLMs) are extremely useful for modeling out-of-vocabulary words by nature. However, their performance is generally much worse than the word-level language models (WLMs), since CLMs need to consider longer history of tokens to properly predict the next one. We address this problem by proposing hierarchical RNN architectures, which consist of multiple modules with different timescales. Despite the multi-timescale structures, the input and output layers operate with the character-level clock, which allows the existing RNN CLM training approaches to be directly applicable without any modifications. Our CLM models show better perplexity than Kneser-Ney (KN) 5-gram WLMs on the One Billion Word Benchmark with only 2\% of parameters. Also, we present real-time character-level end-to-end speech recognition examples on the Wall Street Journal (WSJ) corpus, where replacing traditional mono-clock RNN CLMs with the proposed models results in better recognition accuracies even though the number of parameters are reduced to 30\%.
\end{abstract}
\section{Introduction}

Language models (LMs) show the probability distribution over sequences of words or characters, and they are very important for many speech and document processing applications including speech recognition, text generation, and machine translation \cite{rabiner1993fundamentals, sutskever2011generating, brown1990statistical}. LMs can be classified into character-, word-, and context-levels according to the unit of the input and output. In the character-level LM (CLM) \cite{sutskever2011generating}, the probability distribution of the next characters are generated based on the past character sequences.  Since the number of alphabets is small in English, for example, the input and output of the CLM is quite simple. However, the word-level LM (WLM) is usually needed because the character-level modeling is disadvantaged in utilizing the long period of past sequences.  However, the problem of the word-level model is the complexity of the input and output because the vocabulary size to be supported can be bigger than 1 million. 

LMs have long been developed by analyzing a large amount of texts and storing the probability distribution of word sequences into the memory. The statistical language model demands a large memory space, often exceeding 1 GB, not only because the vocabulary size is large but also their combinations needs to be considered.  In recent years, the language modeling based on recurrent neural networks (RNNs) have been actively investigated \cite{mikolov2010recurrent, jozefowicz2016exploring}. However, the RNN based WLMs still demand billions of parameters because of the large vocabulary size.

In this work, we propose hierarchical RNN based LMs that combine the advantageous characteristics of both character- and word-level LMs.  The proposed network consists of a low-level and a high-level RNNs.  The low-level RNN employs the character-level input and output, and provides the short-term embedding to the high-level RNN that operates as the word-level RNN.  The high-level RNN do not need complex input and output because it receives the character-embedding information from the low-level network, and sends the word-prediction information back to the low-level in a compressed form.  Thus, when considering the input and output, the proposed network is a CLM, although it contains a word-level model inside. The low-level module operates with the character input clock, while the high-level one runs with the space (\texttt{<w>}) and sentence boundary tokens (\texttt{<s>}) that separates words. We expect this hierarchical LM can be extended for processing a longer period of information, such as sentences, topics, or other contexts.% The proposed hierarchical LM can be trained with the character based texts, such as Wikipedia or Wall Street Journal (WSJ) corpus \cite{paul1992design}.

%This paper is organized as follows. Section~2 describes the related work, and Section~3 explains the character-level language modeling using RNNs. RNN modeling including the external clock and reset signals is shown in Section~4, and the proposed language model using a hierarchical RNN is presented in Section~5.  Section~6 gives the experimental results, and concluding remarks are given in Section~7. 

\section{Related work}

\subsection{Character-level language modeling with RNNs}

CLMs need to consider longer sequence of history tokens to predict the next token than the WLMs, due to the smaller unit of tokens. Therefore, traditional $N$-gram models cannot be employed for CLMs. Thanks to the recent advances in RNNs, RNN-based CLMs has begun to show satisfactory performances \cite{sutskever2011generating, hermans2013training}. Especially, deep long short-term memory (LSTM) \cite{hochreiter1997long} based CLMs show excellent performance and successfully applied to end-to-end speech recognition system \cite{hwang2016character}.

\begin{figure}[t]
  \centering
\includegraphics[width=0.9\linewidth]{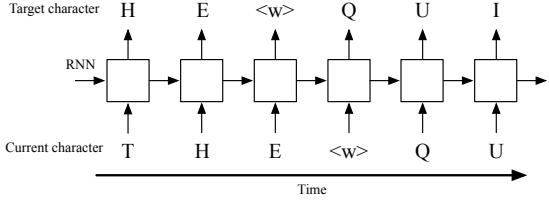}%
  \caption{Training an RNN-based CLM.}
  \label{fig:CLM_train}
\end{figure}

For training RNN CLMs, training data should be first converted to the sequence of one-hot encoded character vectors, $\mathbf x_t$, where the characters include word boundary symbols, \texttt{<w>} or space, and optionally sentence boundary symbols, \texttt{<s>}. Then, as shown in \figurename~\ref{fig:CLM_train}, the RNN is trained to predict the next character $\mathbf x_{t+1}$ by minimizing the cross-entropy loss of the softmax output \cite{bridle1990probabilistic} that represents the probability distributions of the next character.

\subsection{Character-aware word-level language modeling}

There has been many attempts to make WLMs understand character-level inputs. One of the most successful approaches is to encode the arbitrary character sequence to fixed dimensional vector, which is called word embedding, and feed this vector to the word-level RNN LMs. In \cite{kim2015character}, convolutional neural networks (CNNs) are used to generate word embeddings, and achieved the state of the art results on English Penn Treebank corpus \cite{marcus1993building}. The similar CNN-based embedding approach is used by \cite{jozefowicz2016exploring} with very large LSTM networks on the One Billion Word Benchmark \cite{chelba2013one}, also achieving the state of the art perplexity. In \cite{ling2015finding, miyamoto2016gated}, bidirectional LSTMs are employed instead of CNNs for word embedding. However, in all of these approaches, LMs still generate the output probabilities at the word-level. Although the character-level modeling approach of the output word probability is introduced using CNN softmax in \cite{jozefowicz2016exploring}, the base LSTM network still runs with a word-level clock.

Our approach is different from the above ones in many ways. First, our base model is the character-level RNN LMs, instead of WLMs, and we extend this model to enhance the model to consider long-term contexts. Therefore, the output probabilities are generated with a character-level clocks. This property is extremely useful for character-level beam search for end-to-end speech recognition \cite{hwang2016character}. Also, the input and output of our model are the same as those of the traditional character-level RNNs, thus the same training algorithm and recipe can be used without any modifications. Furthermore, the proposed models have significantly less number of parameters compared to WLM-based ones, since the size of our model does not directly depend on the vocabulary size of the training set. Note that a similar hierarchical concept has been used for character-level machine translation \cite{ling2015character}. However, we propose more general hierarchical unidirectional RNN architecture that can be applied for various applications.

\section{RNNs with external clock and reset signals}

In this section, we generalize the existing RNN structures and extend them with external clocks and reset signals. The extended models become the basic building blocks of the hierarchical RNNs.

Most types of RNNs or recurrent layers can be generalized as
\begin{align}
\mathbf s_t = f( \mathbf x_t,  \mathbf s_{t-1}) \;,\;\;
\mathbf y_t = g( \mathbf s_t)
\end{align}
where $\mathbf x_t$ is the input, $\mathbf s_t$ is the state, $\mathbf y_t$ is the output at time step $t$, $f(\cdot)$ is the recurrence function, and $g(\cdot)$ is the output function. For example, a hidden layer of Elman networks \cite{elman1990finding} can be written as
\begin{align}
\mathbf y_t = \mathbf s_t = \mathbf h_t = \sigma ( W_{hx} \mathbf x_t + W_{hh} \mathbf h_{t-1} + \mathbf b_h )
\end{align}
where $\mathbf h_t$ is the activation of the hidden layer, $\sigma(\cdot)$ is the activation function, $W_{hx}$ and $W_{hh}$ are the weight matrices and $\mathbf b_h$ is the bias vector.

LSTMs \cite{hochreiter1997long} with forget gates \cite{gers2000learning} and peephole connections \cite{gers2003learning} can also be converted to the generalized form. The forward equations of the LSTM layer are as follows:
\begin{align}
\mathbf i_t &= \sigma(W_{ix} \mathbf x_t + W_{ih} \mathbf h_{t-1} + W_{im} \mathbf m_{t-1} + \mathbf b_i) \\
\mathbf f_t &= \sigma(W_{fx} \mathbf x_t + W_{fh} \mathbf h_{t-1} + W_{fm} \mathbf m_{t-1} + \mathbf b_f) \\
\mathbf m_t &= \mathbf f_t \circ \mathbf m_{t-1} + \mathbf i_t \circ \tanh ( W_{mx} \mathbf x_t + W_{mh} \mathbf h_{t-1} + \mathbf b_m) \\
\mathbf o_t &= \sigma(W_{ox} \mathbf x_t + W_{oh} \mathbf h_{t-1} + W_{om} \mathbf m_{t} + \mathbf b_o) \\
\mathbf h_t &= \mathbf o_t \circ \tanh(\mathbf m_t)
\end{align}
where $\mathbf i_t$, $\mathbf f_t$, and $\mathbf o_t$ are the input, forget, and output gate values, respectively, $\mathbf m_t$ is the memory cell state, $\mathbf h_t$ is the output activation of the layer, $\sigma(\cdot)$ is the logistic sigmoid function, and $\circ$ is the element-wise multiplication operator. These equations can be generalized by setting $\mathbf s_t = [ \mathbf m_t, \mathbf h_t ]$ and $\mathbf y_t = \mathbf h_t$.

Any generalized RNNs can be converted to the ones that incorporate an external clock signal, $c_t$, as
\begin{align}
\mathbf s_t = (1-c_t) \mathbf s_{t-1} + c_t f( \mathbf x_t,  \mathbf s_{t-1}) \label{eq:clocked} \;,\;\; \mathbf y_t = g( \mathbf s_t)
\end{align}
where $c_t$ is 0 or 1. The RNN updates its state and output only when $c_t=1$. Otherwise, when $c_t=0$, the state and output values remain the same as those of the previous step.

 The reset of RNNs is performed by setting $\mathbf s_{t-1}$ to 0. Specifically, \eqref{eq:clocked} becomes
 \begin{align}
\mathbf s_t &= (1-c_t) (1 - r_t) \mathbf s_{t-1} + c_t f( \mathbf x_t,  (1 - r_t) \mathbf s_{t-1}) \label{eq:reset}
\end{align}
where the reset signal $r_t = 0$ or $1$. When $r_t = 1$, the RNN forgets the previous contexts.

If the original RNN equations are differentiable, the extended equations with clock and reset signals are also differentiable. Therefore, the existing gradient-based training algorithms for RNNs, such as backpropagation through time (BPTT), can be employed for training the extended versions without any modifications.

\section{Character-level language modeling with a hierarchical RNN}

\begin{figure}[t]
  \centering
  \includegraphics[width=1\linewidth]{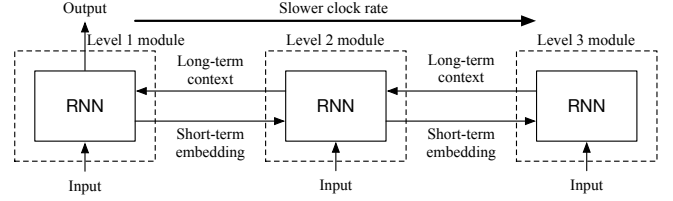}%
  \caption{Hierarchical RNN (HRNN).}
  \label{fig:HRNN}
\end{figure}

The proposed hierarchical RNN (HRNN) architectures have several RNN modules with different clock rates as depicted in \figurename~\ref{fig:HRNN}. The higher level module employs a slower clock rate than the lower module, and the lower level module is reset at every clock of the higher level module. Specifically, if there are $L$ hierarchy levels, then the RNN consists of $L$ submodules. Each submodule $l$ operates with an external clock $c_{l, t}$ and a reset signal $r_{l, t}$, where $l=1,\cdots,L$. The lowest level module, $l=1$, has the fastest clock rate, that is, $c_{1, t}=1$ for all $t$. On the other hand, the higher level modules, $l>1$, have slower clock rates and $c_{l, t}$ can be 1 only when $c_{l-1, t}$ is 1. Also, the lower level modules $l < L$ are reset by the higher level clock signals, that is, $r_{l, t} = c_{l+1, t}$.

The hidden activations of a module, $l < L$, are fed to the next higher level module, $l+1$, \emph{delayed by one time step} to avoid unwanted reset by $r_{l, t} = c_{l+1, t}=1$. This hidden activation vector, or embedding vector, contains compressed short-term context information. The reset of the module by the higher level clock signals helps the module to concentrate on compressing only the short term information, rather than considering longer dependencies. The next higher level module, $l + 1$, process this short-term information to generate the long-term context vector, which is fed back to the lower level module, $l$. There is no delay for this context propagation.

\begin{figure}[t]
  \centering
  \subfloat[][HLSTM-A]{\includegraphics[width=0.48\linewidth]{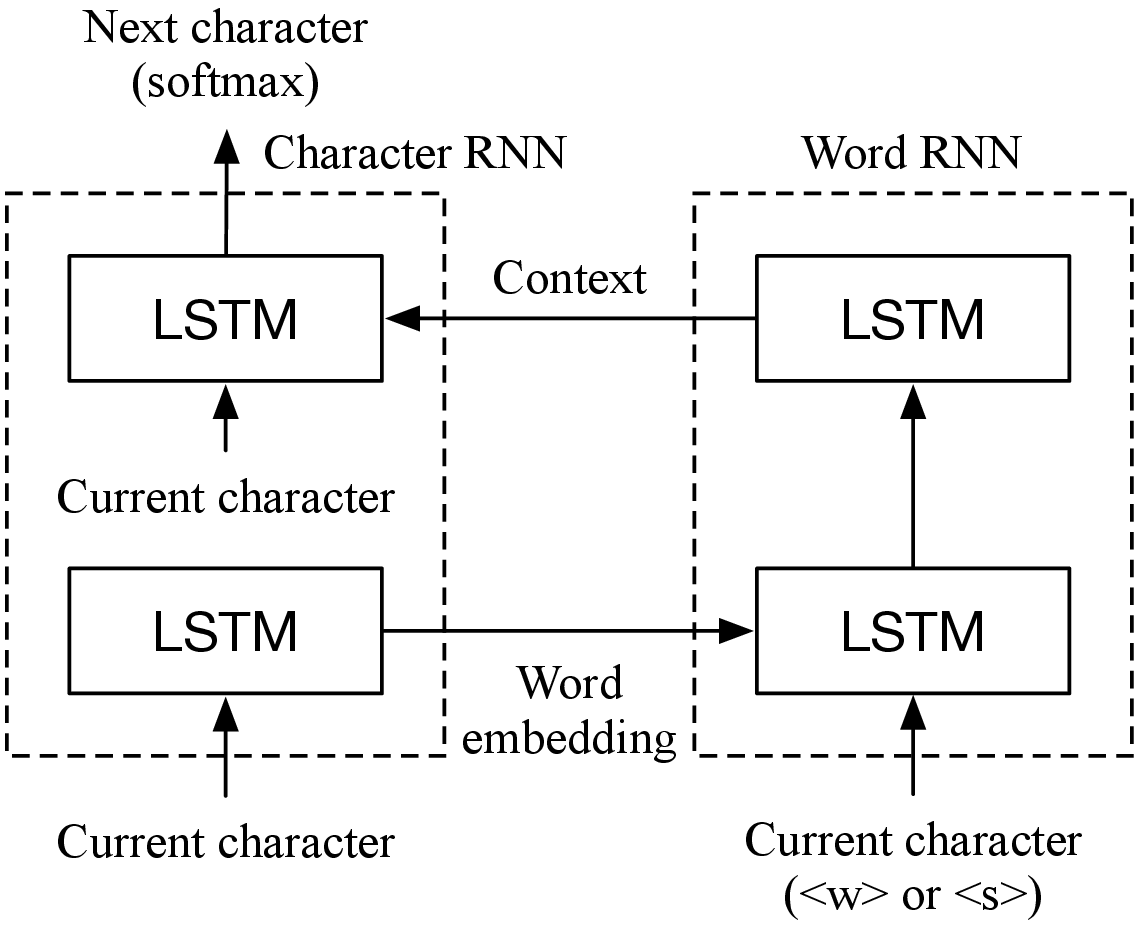}\label{fig:HLSTM-A}}%
  \hspace{0.03\linewidth}%
  \subfloat[][HLSTM-B]{\includegraphics[width=0.48\linewidth]{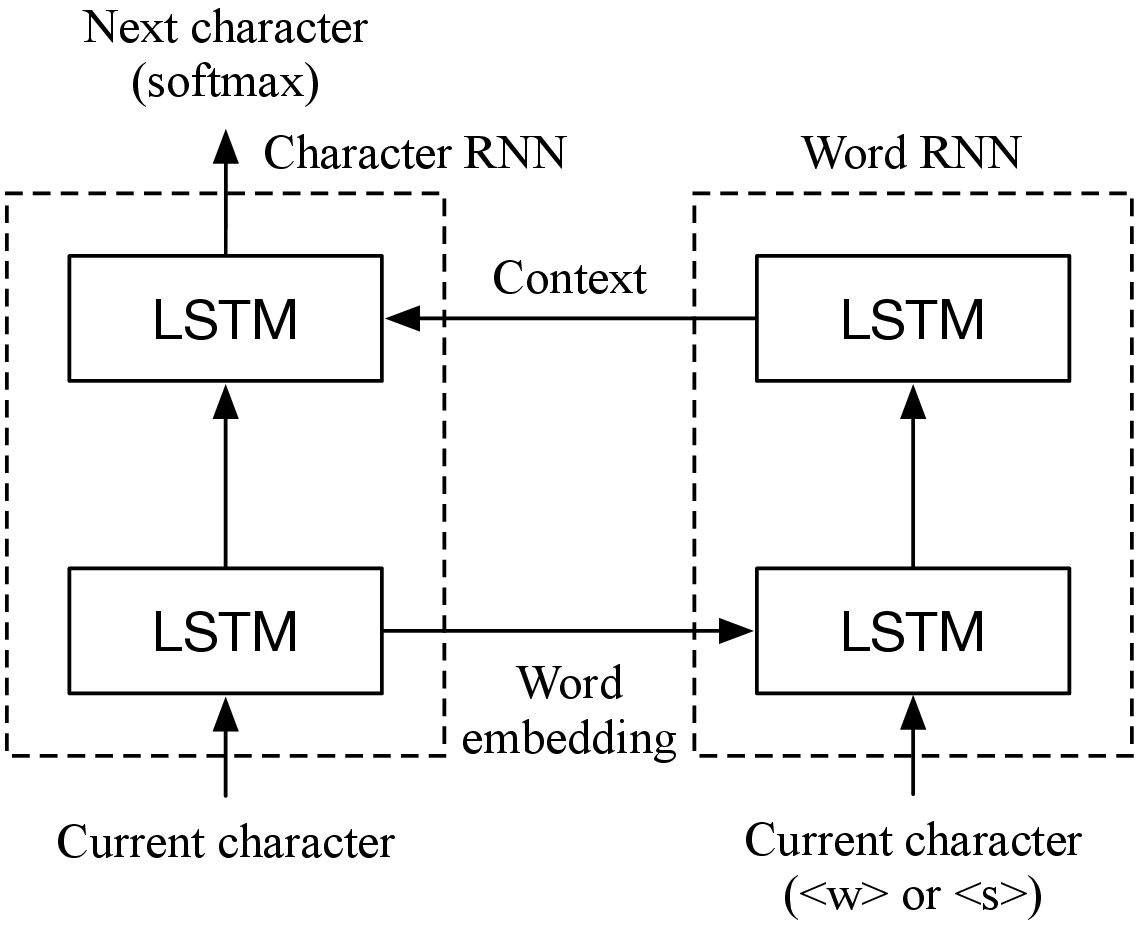}\label{fig:HLSTM-B}}%
  \caption{Two-level hierarchical LSTM (HLSTM) structures for CLMs.}
  \label{fig:HLSTM}
\end{figure}

For character-level language modeling, we use a two-level ($L=2$) HRNN with letting $l=1$ be a character-level module and $l=2$ be a word-level module. The word-level module is clocked at the word boundary input, \texttt{<w>}, which is usually a whitespace character. The input and softmax output layer is connected to the character-level module, and the current word boundary token (e.g. \texttt{<w>} or \texttt{<s>}) information is given to the word-level module. Since this HRNNs have a scalable architecture, we expect this HRNN CLM can be extended for modeling sentence-level contexts by adding an additional sentence-level module, $l=3$. In this case, the sentence-level clock, $c_{3, t}$ becomes 1 when the input character is a sentence boundary token \texttt{<s>}. Also, the word-level module should be clocked at both the word boundary input, \texttt{<w>}, and the sentence boundary input, \texttt{<s>}. In this paper, the experiments are performed only with the two-level HRNN CLMs.

We propose two types of two-level HRNN CLM architectures. As shown in \figurename~\ref{fig:HLSTM}, both models have two LSTM layers per submodule. Note that each connection has a weight matrix. In the HLSTM-A architecture, both LSTM layers in the character-level module receives one-hot encoded character input. Therefore, the second layer of the character-level module is a generative model conditioned by the context vector. On the other hand, in HLSTM-B, the second LSTM layer of the character-level module does not have direct connection from the character inputs. Instead, a word embedding from the first LSTM layer is fed to the second LSTM layer, which makes the first and second layers of the character-level module work together to estimate the next character probabilities when the context vector is given. The experimental results show that HLSTM-B is more efficient for CLM applications.

Since the character-level modules are reset by the word-boundary token (i.e. \texttt{<w>} or whitespace), the context vector from the word-level module is the only source for the inter-word context information. Therefore, the model is trained to generate the context vector that contains useful information about the probability distribution of the next word. From this perspective, the word-level module in both HRNN CLM architectures can be considered as a word-level RNN LM, where the input is a word embedding vector and the output is a compressed descriptor of the next word probabilities. Although the proposed model consists of several RNN modules with different timescales, these can be jointly trained by BPTT as described in Section~3.

\section{Experiments}

The proposed HRNN based CLMs are evaluated with two text datasets: the Wall Street Journal (WSJ) corpus \cite{paul1992design} and One Billion Word Benchmark \cite{chelba2013one}. Also, we present an end-to-end speech recognition example, where HLSTM CLMs are employed for prefix tree-based beam search decoding.

The RNNs are trained with truncated backpropagation through time (BPTT) \cite{werbos1990backpropagation, williams1990efficient}. Also, ADADELTA \cite{zeiler2012adadelta} and Nesterov momentum \cite{nesterov1983method} is applied for weight update. No regularization method, such as dropout \cite{hinton2012improving}, is employed. The training is accelerated using GPUs by training multiple sequences in parallel \cite{hwang2015single}.

\subsection{Perplexity}

In this section, our models are compared with other WLMs in the literature in terms of word-level perplexity (PPL). The word-level PPL of our models is directly converted from bits-per-character (BPC), which is the standard performance measure for CLMs, as follows:
\begin{align}
	PPL = 2^{BPC \times \frac{ N_c}{N_w}}
\end{align}
where $N_c$ and $N_w$ are the number of characters and words in a test set, respectively. Note that sentence boundary symbols (\texttt{<s>}) are also regarded as characters and words.

\subsubsection{Wall Street Journal (WSJ) corpus}
\label{sssec:PPL_WSJ}

\begin{table}[t]
  \caption{Perplexities of CLMs on the WSJ corpus}
  \label{tbl:WSJ_CLM}
  \centering
  \begin{tabular}{lllll}
    \toprule
    Model & Size & \# Params & BPC & Word PPL \\
    \midrule %5.7798680429 chars per word
    Deep LSTM & 2x512 & 3.23 M & 1.148 & 99.5 \\ % 1.148251
    Deep LSTM & 4x512 & 7.43 M & 1.132 & 93.3 \\ % 1.132206
    Deep LSTM & 4x1024 & 29.54 M & 1.101 & 82.4 \\ % 1.101270
    \midrule
    HLSTM-A & 4x512 & 7.50 M & 1.089 & 78.5 \\ % 1.089111
    HLSTM-B (no reset) & 4x512 & 8.48 M & 1.080 & 75.7 \\ % 1.079935
    HLSTM-B & 4x512 & 8.48 M & 1.073 & 73.6 \\ % 1.072868
    HLSTM-B & 4x1024 & 33.74 M & 1.058 & 69.2 \\ % 1.057583
    \bottomrule
  \end{tabular}
\end{table}

\begin{table}[t]
  \caption{Perplexities of WLMs on the WSJ corpus in the literature}
  \label{tbl:WSJ_WLM}
  \centering
  \begin{tabular}{lllll}
    \toprule
    Model & \# Params & PPL \\
    \midrule
    KN 5-gram (no count cutoffs) \cite{mikolov2012statistical} & -  & 80 \\
    RNN-640 + ME 4-gram feature \cite{mikolov2012statistical}  & 2 G  & 59 \\
    \bottomrule
  \end{tabular}
\end{table}

%\paragraph{Dataset}
The Wall Street Journal (WSJ) corpus \cite{paul1992design} is designed for training and benchmarking automatic speech recognition systems. For the perplexity experiments, we used the non-verbalized punctuation (NVP) version of the LM training data inside the corpus. The dataset consists of about 37 million words, where one percent of the total data is held out for the final evaluation and does not participate in training. All alphabets are converted to the uppercases.

%\paragraph{Experimental results}
\tablename~\ref{tbl:WSJ_CLM} shows the perplexities of traditional mono-clock deep LSTM and HLSTM based CLMs on the held-out set. Note that the size $N$x$M$ means that the network consists of $N$ LSTM layers, where each layer contains $M$ memory cells. The HLSTM models show better perplexity performanes even when the number of LSTM cells or parameters is much smaller than that of the deep LSTM networks. Especially, HLSTM-B network with the size of 4x512 has about 9\% lower perplexity than deep LSTM (4x1024) model, even with only 29\% of parameters.

%\paragraph{Importance of reset}
It is important to reset the character-level modules at the word-level clocks for helping the character-level modules to better concentrate on the short-term information. As observed in \tablename~\ref{tbl:WSJ_CLM}, removing the reset functionality of the character-level module of the HLSTM-B model results in degraded performance.

%\paragraph{Comparison with WLMs}
The non-ensemble perplexities of WLMs in the literature are presented in \tablename~\ref{tbl:WSJ_WLM}. The Kneser-Ney (KN) smoothed 5-gram model (KN-5) \cite{kneser1995improved} is a strong non-neural WLM baseline. With the standard deep RNN based CLMs, it is very hard to beat KN-5 in terms of perplexity. However, it is surprising that all HLSTM models in \tablename~\ref{tbl:WSJ_CLM} shows better perplexities than KN-5 does. The RNN based WLM model combined with the maximum entropy 4-gram feature \cite{mikolov2012context, mikolov2012statistical} shows much better results than the proposed HLSTM based CLM models. However, like most of the WLMs, it also needs a very large number (2~G) of parameters and cannot handle out-of-vocabulary (OOV) words.

\subsubsection{One Billion Word Benchmark}

\begin{table}[t]
  \caption{Perplexities of the HRNN CLMs on the One Billion Word Benchmark}
  \label{tbl:1B_CLM}
  \centering
  \begin{tabular}{lllll}
    \toprule %CPW: 5.1954364955
    Model & Size & \# Params & BPC & Word PPL \\
    \midrule
    HLSTM-B & 4x512 & 9.06 M & 1.228 & 83.3 \\ %1.228207
    HLSTM-B & 4x1024 & 34.90 M & 1.140 & 60.7 \\ %1.140078
    \bottomrule
  \end{tabular}
\end{table}

\begin{table}[t]
  \caption{Perplexities of WLMs on the One Billion Word Benchmark in the literature}
  \label{tbl:1B_WLM}
  \centering
  \begin{tabular}{lllll}
    \toprule
    Model & \# Params & PPL \\
    \midrule
    Sigmoid RNN-2048 \cite{ji2015blackout}  & 4.1 G & 68.3 \\
    Interpolated KN-5, 1.1B n-grams \cite{chelba2013one}  & 1.76 G & 67.6 \\
    LightRNN \cite{li2016lightrnn} & \bf{41 M} & \bf{66} \\
    Sparse non-negative matrix LM \cite{shazeer2015sparse}  & 33 G & 52.9 \\
    RNN-1024 + ME 9-gram feature \cite{chelba2013one}  & 20 G & 51.3 \\
    CNN input + 2xLSTM-8192-1024 \cite{jozefowicz2016exploring}  & 1.04 G & 30.0 \\
    \bottomrule
  \end{tabular}
\end{table}

%\paragraph{Dataset}

The One Billion Word Benchmark \cite{chelba2013one} dataset contains about 0.8 billion words and roughly 800 thousand words of vocabulary. We followed the standard way of splitting the training and test data as in \cite{chelba2013one}. Each byte of UTF-8 encoded text is regarded as a character. Therefore, the size of the character set is 256.

%\paragraph{Experimental results}

Due to the large amount of training data and weeks of training time, only two HLSTM-B experiments are conducted with the size of 4x512 and 4x1024. As shown in \tablename~\ref{tbl:1B_CLM}, there are large gap (22.5) in word-level perplexity between the two models. Therefore, further improvement in perplexity can be expected with bigger networks.

%\paragraph{Comparison with WLMs}

The perplexities of other WLMs are summarized in \tablename~\ref{tbl:1B_WLM}. The proposed HLSTM-B model (4x1024) shows better perplexities than the interpolated KN-5 model with 1.1 billion n-grams \cite{chelba2013one} even though the number of parameters of our model is only 2\% of that of the KN-5 model. Also, our model performs better than LightRNN \cite{li2016lightrnn}, which is a word-level RNN LM that has about 17\% more parameters than ours. However, much lower perplexities are reported with sparse non-negative matrix LM and the maximum entropy feature based RNN model \cite{chelba2013one}, where the number of parameters are 33 G and 20 G, respectively. Recently, the state of the art perplexity of 30.0 was reported in \cite{jozefowicz2016exploring} with a single model that has 1 G parameters. The model is basically a very large LSTM LM. However, the convolutional neural network (CNN) is used to generate word embedding of arbitrary character sequences as the input of the LSTM LM. Therefore, this model can handle OOV word inputs, however, still the model runs with a word-level clock.

\subsection{End-to-end automatic speech recognition (ASR)}

\begin{table}[t]
  \caption{End-to-end ASR results on the WSJ Nov'92 20K evaluation set (\texttt{eval92})}
  \label{tbl:ASR}
  \centering
  \begin{tabular}{llllll}
    \toprule
    Model & Size & \# Params & Word PPL & WER  \\
    \midrule
    Deep LSTM & 4x512 & 7.43 M & 93.3 & 8.36\% \\%& 6.01\% \\
    Deep LSTM & 4x1024 & 29.54 M & 82.4 & 7.85\% \\%& 5.48\% \\
    \midrule
    HLSTM-B & 4x512 & 8.48 M & 73.6 & 7.79\% \\%& 5.49\% \\
    HLSTM-B & 4x1024 & 33.74 M & 69.2 & 7.78\% \\%& 5.68\%\\
    \bottomrule
  \end{tabular}
\end{table}

In this section, we apply the proposed CLMs to the end-to-end automatic speech recognition (ASR) system to evaluate the models in more practical situation than just measuring perplexities. The CLMs are trained with WSJ LM training data as in Section~\ref{sssec:PPL_WSJ}. Unlike WLMs, the proposed CLMs have very small number of parameters, so they can be employed for real-time character-level beam search. %All the experiments in this section run in real-time with NVIDIA GeForce GTX Titan X GPU.

The incremental speech recognition system proposed in \cite{hwang2016character} is used for the evaluation. The acoustic model is 4x512 unidirectional LSTM and end-to-end trained with connectionist temporal classification (CTC) loss \cite{graves2006connectionist} using the non-verbalized punctuation (NVP) portion of WSJ SI-284 training set. The acoustic features are 40-dimensional log-mel filterbank coefficients, energy and their delta and double-delta values, which are extracted every 10 ms with 25 ms Hamming window. The beam-search decoding is performed on a prefix-tree with depth-pruning and width-pruning \cite{hwang2016character}. The insertion bonus is 1.6, the LM weight is 2.0, and the beam width is 512.

The results are summarized in \tablename~\ref{tbl:ASR}. It is observed that the perplexity of LM and the word error rate (WER) have strong correlation. As shown in the table, we can achieve a better WER by replacing the traditional deep LSTM (4x1024) CLM with the proposed HLSTM-B (4x512) CLM, while reducing the number of LM parameters to 30\%.

\section{Concluding remarks}

In this paper, hierarchical RNN (HRNN) based CLMs are proposed. The HRNN consists of several submodules with different clock rates. Therefore, it is capable of learning long-term dependencies as well as short-term details. The experimental results on One Billion Benchmark show that HLSTM-B networks significantly outperform Kneser-Ney 5-gram LMs with only 2\% of parameters. Although other RNN-based WLMs show better performance than our models, they have impractically many parameters. On the other hand, as shown in the WSJ speech recognition example, the proposed model can be employed for the real-time speech recognition with less than 10 million parameters. Also, CLMs can handle OOV words by nature, which is a great advantage for the end-to-end speech recognition and many NLP tasks. One of the interesting future work is to train the clock signals, instead of using manually designed ones. %Also, it would be interesting to see how this hierarchical architecture will perform when the level of hierarchy increases.

\bibliographystyle{IEEEbib}
\bibliography{refs}

\begin{thebibliography}{10}

\bibitem{rabiner1993fundamentals}
Lawrence Rabiner and Biing-Hwang Juang,
\newblock {\em Fundamentals of speech recognition},
\newblock Prentice Hall, 1993.

\bibitem{sutskever2011generating}
Ilya Sutskever, James Martens, and Geoffrey~E Hinton,
\newblock ``Generating text with recurrent neural networks,''
\newblock in {\em Proceedings of the 28th International Conference on Machine
  Learning (ICML-11)}, 2011, pp. 1017--1024.

\bibitem{brown1990statistical}
Peter~F Brown, John Cocke, Stephen A~Della Pietra, Vincent J~Della Pietra,
  Fredrick Jelinek, John~D Lafferty, Robert~L Mercer, and Paul~S Roossin,
\newblock ``A statistical approach to machine translation,''
\newblock {\em Computational linguistics}, vol. 16, no. 2, pp. 79--85, 1990.

\bibitem{mikolov2010recurrent}
Tomas Mikolov, Martin Karafi{\'a}t, Lukas Burget, Jan Cernock{\`y}, and Sanjeev
  Khudanpur,
\newblock ``Recurrent neural network based language model.,''
\newblock in {\em Proc. Interspeech}, 2010, vol.~2, p.~3.

\bibitem{jozefowicz2016exploring}
Rafal Jozefowicz, Oriol Vinyals, Mike Schuster, Noam Shazeer, and Yonghui Wu,
\newblock ``Exploring the limits of language modeling,''
\newblock {\em arXiv preprint arXiv:1602.02410}, 2016.

\bibitem{hermans2013training}
Michiel Hermans and Benjamin Schrauwen,
\newblock ``Training and analysing deep recurrent neural networks,''
\newblock in {\em Advances in Neural Information Processing Systems}, 2013, pp.
  190--198.

\bibitem{hochreiter1997long}
Sepp Hochreiter and J{\"u}rgen Schmidhuber,
\newblock ``Long short-term memory,''
\newblock {\em Neural computation}, vol. 9, no. 8, pp. 1735--1780, 1997.

\bibitem{hwang2016character}
Kyuyeon Hwang and Wonyong Sung,
\newblock ``Character-level incremental speech recognition with recurrent
  neural networks,''
\newblock in {\em 2016 IEEE International Conference on Acoustics, Speech and
  Signal Processing (ICASSP)}, 2016.

\bibitem{bridle1990probabilistic}
John~S Bridle,
\newblock ``Probabilistic interpretation of feedforward classification network
  outputs, with relationships to statistical pattern recognition,''
\newblock in {\em Neurocomputing}, pp. 227--236. Springer, 1990.

\bibitem{kim2015character}
Yoon Kim, Yacine Jernite, David Sontag, and Alexander~M Rush,
\newblock ``Character-aware neural language models,''
\newblock in {\em Thirtieth AAAI Conference on Artificial Intelligence}, 2016.

\bibitem{marcus1993building}
Mitchell~P Marcus, Mary~Ann Marcinkiewicz, and Beatrice Santorini,
\newblock ``Building a large annotated corpus of english: The penn treebank,''
\newblock {\em Computational linguistics}, vol. 19, no. 2, pp. 313--330, 1993.

\bibitem{chelba2013one}
Ciprian Chelba, Tomas Mikolov, Mike Schuster, Qi~Ge, Thorsten Brants, Phillipp
  Koehn, and Tony Robinson,
\newblock ``One billion word benchmark for measuring progress in statistical
  language modeling,''
\newblock {\em arXiv preprint arXiv:1312.3005}, 2013.

\bibitem{ling2015finding}
Wang Ling, Tiago Lu{\'\i}s, Lu{\'\i}s Marujo, Ram{\'o}n~Fernandez Astudillo,
  Silvio Amir, Chris Dyer, Alan~W Black, and Isabel Trancoso,
\newblock ``Finding function in form: Compositional character models for open
  vocabulary word representation,''
\newblock in {\em 2015 Conference on Empirical Methods in Natural Language
  Processing}, 2015, pp. 1520--1530.

\bibitem{miyamoto2016gated}
Yasumasa Miyamoto and Kyunghyun Cho,
\newblock ``Gated word-character recurrent language model,''
\newblock in {\em 2016 Conference on Empirical Methods in Natural Language
  Processing}, 2016, pp. 1992--1997.

\bibitem{ling2015character}
Wang Ling, Isabel Trancoso, Chris Dyer, and Alan~W Black,
\newblock ``Character-based neural machine translation,''
\newblock in {\em Proceedings of the 54th Annual Meeting of the Association for
  Computational Linguistics}, 2016, vol. 357--361.

\bibitem{elman1990finding}
Jeffrey~L Elman,
\newblock ``Finding structure in time,''
\newblock {\em Cognitive Science}, vol. 14, no. 2, pp. 179--211, 1990.

\bibitem{gers2000learning}
Felix~A Gers, J{\"u}rgen Schmidhuber, and Fred Cummins,
\newblock ``Learning to forget: Continual prediction with {LSTM},''
\newblock {\em Neural Computation}, vol. 12, no. 10, pp. 2451--2471, 2000.

\bibitem{gers2003learning}
Felix~A Gers, Nicol~N Schraudolph, and J{\"u}rgen Schmidhuber,
\newblock ``Learning precise timing with {LSTM} recurrent networks,''
\newblock {\em The Journal of Machine Learning Research}, vol. 3, pp. 115--143,
  2003.

\bibitem{paul1992design}
Douglas~B Paul and Janet~M Baker,
\newblock ``The design for the {W}all {S}treet {J}ournal-based {CSR} corpus,''
\newblock in {\em Proceedings of the workshop on Speech and Natural Language}.
  Association for Computational Linguistics, 1992, pp. 357--362.

\bibitem{werbos1990backpropagation}
Paul~J Werbos,
\newblock ``Backpropagation through time: what it does and how to do it,''
\newblock {\em Proceedings of the IEEE}, vol. 78, no. 10, pp. 1550--1560, 1990.

\bibitem{williams1990efficient}
Ronald~J Williams and Jing Peng,
\newblock ``An efficient gradient-based algorithm for on-line training of
  recurrent network trajectories,''
\newblock {\em Neural Computation}, vol. 2, no. 4, pp. 490--501, 1990.

\bibitem{zeiler2012adadelta}
Matthew~D Zeiler,
\newblock ``{ADADELTA}: An adaptive learning rate method,''
\newblock {\em arXiv preprint arXiv:1212.5701}, 2012.

\bibitem{nesterov1983method}
Yurii Nesterov,
\newblock ``A method of solving a convex programming problem with convergence
  rate {O} (1/k2),''
\newblock {\em Soviet Mathematics Doklady}, vol. 27, no. 2, pp. 372--376, 1983.

\bibitem{hinton2012improving}
Geoffrey~E Hinton, Nitish Srivastava, Alex Krizhevsky, Ilya Sutskever, and
  Ruslan~R Salakhutdinov,
\newblock ``Improving neural networks by preventing co-adaptation of feature
  detectors,''
\newblock {\em arXiv preprint arXiv:1207.0580}, 2012.

\bibitem{hwang2015single}
Kyuyeon Hwang and Wonyong Sung,
\newblock ``Single stream parallelization of generalized {LSTM}-like {RNN}s on
  a {GPU},''
\newblock in {\em 2015 IEEE International Conference on Acoustics, Speech and
  Signal Processing (ICASSP)}. IEEE, 2015, pp. 1047--1051.

\bibitem{mikolov2012statistical}
Tom{\'a}{\v{s}} Mikolov,
\newblock {\em Statistical language models based on neural networks},
\newblock Ph.D. thesis, Brno University of Technology, 2012.

\bibitem{kneser1995improved}
Reinhard Kneser and Hermann Ney,
\newblock ``Improved backing-off for m-gram language modeling,''
\newblock in {\em 1995 International Conference on Acoustics, Speech, and
  Signal Processing (ICASSP)}. IEEE, 1995, vol.~1, pp. 181--184.

\bibitem{mikolov2012context}
Tomas Mikolov and Geoffrey Zweig,
\newblock ``Context dependent recurrent neural network language model,''
\newblock in {\em 2012 IEEE Spoken Language Technology Workshop}, 2012, pp.
  234--239.

\bibitem{ji2015blackout}
Shihao Ji, SVN Vishwanathan, Nadathur Satish, Michael~J Anderson, and Pradeep
  Dubey,
\newblock ``Black{O}ut: Speeding up recurrent neural network language models
  with very large vocabularies,''
\newblock in {\em 4th International Conference on Learning Representations},
  2016.

\bibitem{li2016lightrnn}
Xiang Li, Tao Qin, Jian Yang, Xiaolin Hu, and Tieyan Liu,
\newblock ``{LightRNN}: Memory and computation-efficient recurrent neural
  networks,''
\newblock in {\em Advances in Neural Information Processing Systems}, 2016, pp.
  4385--4393.

\bibitem{shazeer2015sparse}
Noam Shazeer, Joris Pelemans, and Ciprian Chelba,
\newblock ``Sparse non-negative matrix language modeling for skip-grams,''
\newblock in {\em Proc. Interspeech}, 2015, pp. 1428--1432.

\bibitem{graves2006connectionist}
Alex Graves, Santiago Fern{\'a}ndez, Faustino Gomez, and J{\"u}rgen
  Schmidhuber,
\newblock ``Connectionist temporal classification: labelling unsegmented
  sequence data with recurrent neural networks,''
\newblock in {\em Proceedings of the 23rd International Conference on Machine
  Learning}. ACM, 2006, pp. 369--376.

\end{thebibliography}

\end{document}